\documentclass[11pt]{article}

\usepackage{acl}

\usepackage{times}
\usepackage{latexsym}

\usepackage[T1]{fontenc}

\usepackage[utf8]{inputenc}

\usepackage{microtype}

\usepackage{hyperref}
\usepackage{amsfonts}
\usepackage{amsmath}
\usepackage{amsthm} 
\usepackage{mathrsfs}

\usepackage{multirow}
\usepackage{booktabs}
\usepackage[caption=false]{subfig}
\usepackage{xcolor}
\usepackage{tabularx}
\usepackage{widetable}
\usepackage{graphicx}
\graphicspath{{figures/}}

\definecolor{red}{rgb}{1,0,0}

\title{Fine-tuning Happens in Tiny Subspaces: Exploring Intrinsic Task-specific Subspaces of Pre-trained Language Models}

\author{
    Zhong Zhang\textsuperscript{\rm 1,2},
    Bang Liu\textsuperscript{\rm 3,}\begin{NoHyper}\thanks{\;\; Canada CIFAR AI Chair.}\end{NoHyper}\, \textsuperscript{\rm ,}\footnotemark[2] ,
    Junming Shao\textsuperscript{\rm 1,2,}\begin{NoHyper}\thanks{\;\; Corresponding authors.}\end{NoHyper}  \\
    \textsuperscript{\rm 1}University of Electronic Science and Technology of China, Chengdu, China\\
    \textsuperscript{\rm 2}Shenzhen Institute for Advanced Study, UESTC, Shenzhen, China\\
    \textsuperscript{\rm 3}Mila \& Universit{\'e} de Montr{\'e}al, Montr{\'e}al, Canada\\
    \texttt{zhongzhang@std.uestc.edu.cn}, \\ 
    \texttt{bang.liu@umontreal.ca}, \\ 
    \texttt{junmshao@uestc.edu.cn}
}

\begin{document}
\maketitle

\begin{abstract}
Pre-trained language models (PLMs) are known to be overly parameterized and have significant redundancy, indicating a small degree of freedom of the PLMs. Motivated by the observation, in this paper, we study the problem of re-parameterizing and fine-tuning PLMs from a new perspective: Discovery of intrinsic task-specific subspace. Specifically, by exploiting the dynamics of the fine-tuning process for a given task, the parameter optimization trajectory is learned to uncover its intrinsic task-specific subspace. A key finding is that PLMs can be effectively fine-tuned in the subspace with a small number of free parameters. Beyond, we observe some outlier dimensions emerging during fine-tuning in the subspace. Disabling these dimensions degrades the model performance significantly. This suggests that these dimensions are crucial to induce task-specific knowledge to downstream tasks.
\end{abstract}

\section{Introduction}
\label{sec:intro}
Pre-trained Language Models (PLMs) have become the de facto methods for various natural language processing (NLP) tasks \citep{devlin2019bert,radford2019language,liu2020roberta}. The typical paradigm is to pre-train a big language model on large-scale corpora and then fine-tune the model on small task-specific datasets to adapt to the downstream tasks. Despite the great success of this paradigm, two questions still come to our mind: (1) Why can a PLM with hundreds of millions of parameters be successfully fine-tuned on different downstream tasks using only hundreds or thousands of labeled samples? (2) Do we really need a full fine-tuning of all parameters of a PLM to reach state-of-the-art performance on downstream tasks? In this paper, we try to provide a new viewpoint on the two questions, and claim that: \textbf{Fine-tuning happens only in some tiny task-specific subspaces, which can be effectively learned with a small number of free parameters}.

Recent studies have shown that PLMs are highly over-parameterized and robust to pruning \citep{DBLP:conf/iclr/FrankleC19,DBLP:conf/nips/ChenFC0ZWC20,DBLP:conf/emnlp/PrasannaRR20,gordon-etal-2020-compressing,DBLP:conf/acl/LiangZCJLHZC20,DBLP:conf/iclr/LiangJZH0GCZ22}, and can be fine-tuned in parameter-efficient ways \citep{DBLP:conf/acl/GongHSLC0W022,DBLP:conf/acl/ZakenGR22,DBLP:conf/nips/MahabadiHR21,DBLP:conf/acl/LiL20}. This emerging empirical evidence tends to point to one fact that there exist some intrinsic structures in PLMs that are responsible for inducing task-specific knowledge to downstream tasks. Notably, the recent work \citep{DBLP:conf/acl/AghajanyanGZ20} provides a promising conclusion that PLMs can be re-parameterized and fine-tuned in random low-dimensional subspaces using random projection, and the dimensionality of the random subspace is orders of magnitude smaller than the dimensionality of the full parameter space. Their findings implicitly suggest the existence of such intrinsic structure in the PLMs, which is, however, understudied. To bridge this gap, we explicitly demonstrate that there exist task-specific low-dimensional subspaces in which PLMs can be effectively fine-tuned.

Inspired by the low dimensional landscape hypothesis \citep{li2022low} that a training trajectory of a neural network lies in a low-dimensional subspace, in this work, we thus resort to the fine-tuning trajectory to study the intrinsic task-specific subspaces of PLMs. We show that it is possible to uncover the intrinsic task-specific subspaces with a fine-tuning trajectory by finding its principal directions. The uncovered intrinsic task-specific subspaces usually have very low dimensionalities, but are quite effective in inducing task-specific knowledge. For example, by re-parameterizing the encoder and optimizing only 32 free parameters per-layer in the intrinsic task-specific subspace, the model allows achieving nearly the same performance as fine-tuning in the full parameter space. Moreover, we further show that the uncovered intrinsic task-specific subspaces have a certain transferability. 

Beyond this, we find that the model contains some outlier dimensions with abnormal spikes when fine-tuning in the intrinsic task-specific subspaces instead of a random subspace. Disabling these outlier dimensions degrades the model performance significantly. We believe that this phenomenon is related to the previously discovered outlier dimensions of PLMs \citep{DBLP:conf/acl/LuoKM20,DBLP:conf/acl/KovalevaKRR21,puccetti2022outliers}. However, there are essential differences between them, which we will discuss in the latter section.

By exploring the intrinsic task-specific subspaces of PLMs, the main contributions of this paper are summarized as follows.
\begin{enumerate}
  \item  We interpret the ease of adapting PLMs to downstream tasks as fine-tuning happens in tiny intrinsic task-specific subspaces. Within this interpretation, we propose a method to uncover the subspaces by finding the principal directions of the fine-tuning trajectory.
  \item  We conduct extensive experiments on the GLUE benchmark using BERT and RoBERTa models to support our claims. We show that the models can be effectively fine-tuned with a very small number of parameters in the uncovered intrinsic task-specific subspaces.
  \item  We identify some outlier dimensions when fine-tuning in the intrinsic task-specific subspaces, and some empirical analysis is further given.
\end{enumerate}

\section{Related Work}
\label{sec:rw}
\noindent \textbf{Intrinsic Dimensionality.} 
\citet{DBLP:conf/iclr/LiFLY18} first defined the intrinsic dimension of an objective function in the context of deep learning. They showed that various neural networks can be effectively re-parameterized and trained in random low-dimensional subspaces. Their findings shed light on understanding the high-dimensional landscape of complex neural networks. Following this, \citet{DBLP:conf/acl/AghajanyanGZ20} further measured the intrinsic dimensions of PLMs fine-tuning on downstream tasks. They showed that PLMs have very low intrinsic dimensions ranging from hundreds to thousands. \citet{qin2021exploring} exploited the idea of intrinsic subspace and proposed a prompt tuning method for efficient training. In addition, the concept of intrinsic dimension is also related to the low-rank approximation of PLMs \citep{DBLP:conf/iclr/HuSWALWWC22,DBLP:conf/nips/MahabadiHR21,DBLP:conf/nips/ChenYDH21}, but their motivations are entirely different. The former aims to open the black box of models and explore the internal mechanisms of why they are effective, while the latter focuses on developing new methods to train the models efficiently.

\vspace{1mm}
\noindent \textbf{Random Projection and Subspace Learning.} 
The random projection has a long history in machine learning research community, and is a key tool to analyze the intrinsic dimension \citep{DBLP:conf/iclr/LiFLY18,DBLP:conf/acl/AghajanyanGZ20}. In the context of optimization, \citet{DBLP:conf/nips/GressmannEL20} proposed a random bases descent algorithm to train neural networks in low-dimensional subspaces. However, the random projection inevitably introduces task-irrelevant information, and is not optimal for subspace learning. We believe that a more compact and task-specific subspace can be found in the model, which is the main motivation of this work. \citet{gur2018gradient} empirically found that gradient descent of neural networks happens in a tiny subspace, \citet{li2022low} further developed a subspace learning algorithm DLDR that dynamically extracts the subspace from the optimization trajectory. \citet{DBLP:conf/cvpr/LiWC0H22} leveraged the DLDR algorithm for adversarial training. \citet{pmlr-v199-gauch22a} proposed a SubGD method that identifies the subspace via the eigendecomposition of an auto-correlation matrix for few-shot learning. However, to the best of our knowledge, there is currently no research on the discovery of non-random intrinsic task-specific subspace of PLMs. 

\vspace{1mm}
\noindent \textbf{Outlier Dimensions in Pre-trained Language Models.} 
Multiple studies have identified outlier dimensions in PLMs. Some works were motivated by calibrating the anisotropy behavior of hidden representation of PLMs \citep{DBLP:conf/emnlp/TimkeyS21,ding2021isotropy,DBLP:conf/acl/LuoKM20,su2021whitening,DBLP:conf/emnlp/0004GXMYS20}. Another line of work identified certain outlier dimensions in PLMs that are very sensitive to the fine-tuning of downstream tasks \citep{DBLP:conf/acl/KovalevaKRR21,puccetti2022outliers}. Disabling these outlier dimensions degrades the model performance significantly. \citet{DBLP:conf/acl/LuoKM20} showed that the outlier dimensions are artefacts derived from positional embeddings and layer normalization. \citet{puccetti2022outliers} identified a correlation between outlier dimensions and token frequency. It is worth noting that our findings differ largely from previous works in three ways: 1) The outlier dimensions in their context actually refer to output neurons. In our context, an outlier dimension refers to a specific model parameter. In other words, they consider abnormal outputs, while we consider abnormal weights. 2) The ways of identifying outlier dimensions are different. They identify outlier dimensions by examining abnormal outputs, while we find outlier dimensions by examining abnormal updates to weights. 3) The effects of disabling outlier dimensions are different. They show that disabling just one outlier neuron can result in a significant drop in performance. In contrast, disabling the top outlier weight has almost no effect on the model performance. However, the model performance will drop significantly if we disable more outlier weights. The reason for the emergence of these outlier dimensions remains unclear, and we aim to conduct further in-depth analysis in future work.

\section{Intrinsic Task-specific Subspaces Discovery in PLMs}
\label{sec:main}
\subsection{Preliminary: Intrinsic Dimensionality}
The intrinsic dimension of an objective landscape is first defined by \citet{DBLP:conf/iclr/LiFLY18}, which is the number of independent optimization variables with regard to minimizing the objective function. However, finding the exact intrinsic dimension is computationally intractable for complex objective functions like deep neural networks. Therefore, a random subspace training method is usually employed to estimate the intrinsic dimension \citep{DBLP:conf/iclr/LiFLY18,DBLP:conf/acl/AghajanyanGZ20}.

Formally, let $\boldsymbol{\theta}^D\in \mathbb{R}^D$ be a parameter vector that parameterizes a model $f(\boldsymbol{x} ; \boldsymbol{\theta})$. Take the BERT-base model as an example, $\boldsymbol{\theta}^D$ represents all BERT's parameters that are flattened into a 110M-dimensional vector. $\boldsymbol{\theta}_0^D\in \mathbb{R}^D$ denotes the initial parameterization, $\boldsymbol{P}\in\mathbb{R}^{D\times d}$ denotes a random projection matrix whose columns form an orthonormal basis for a randomly oriented $d$-dimensional subspace of $\mathbb{R}^D$, $\boldsymbol{\theta}^d\in\mathbb{R}^d$ denotes a parameter vector in a lower $d$-dimensional space. The model is fine-tuned in the lower $d$-dimensional subspace via the following re-parameterization method:
\begin{align}
  \boldsymbol{\theta}^D = \boldsymbol{\theta}_0^D +  \boldsymbol{P} \boldsymbol{\theta}^d.
  \label{eq:subspace}
\end{align}
Note that $\boldsymbol{\theta}_0^D$ and $\boldsymbol{P}$ are frozen during the training process, and only $\boldsymbol{\theta}^d$ is trained by the gradient descent. In practice, the re-parameterization can be done in a layer-wise manner to save computational resources \citep{DBLP:conf/acl/AghajanyanGZ20}, and we also follow the layer-wise setting for our analysis.

The intrinsic dimension of a PLM is estimated by grid searching the minimal $d$ that makes the model reach 90\% of the full fine-tuning performance. Take the BERT-base model as an example, the intrinsic dimension for fine-tuning on the MRPC dataset is only 1861 \citep{DBLP:conf/acl/AghajanyanGZ20}, which is surprisingly small considering the original model has up to 110 million parameters.

\begin{figure}[t]
  \begin{center}
  \includegraphics [width=2in]{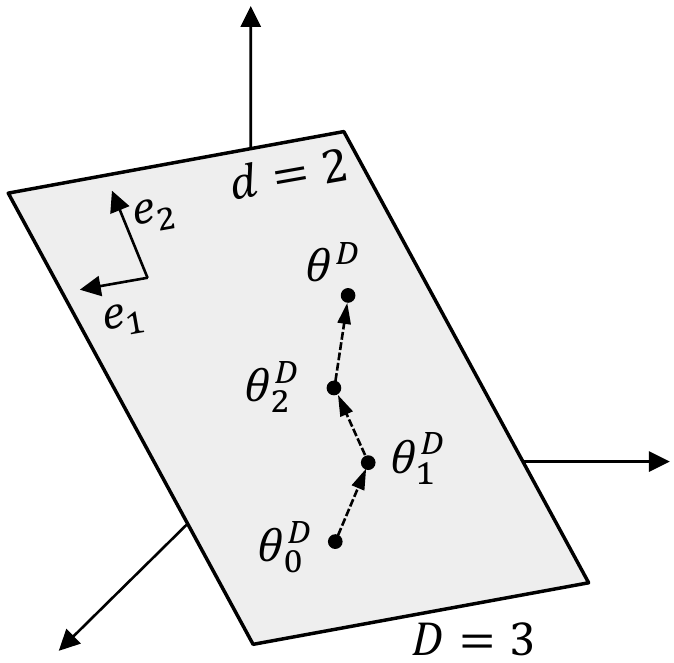}  
  \end{center}
  \caption{An illustrative example of optimizing a model in the 3-dimensional space, while the optimization trajectory only lies in a 2-dimensional subspace. We call the subspace the intrinsic subspace for the model.}
  \label{fig:subspace}
\end{figure}

\subsection{Finding Intrinsic Task-specific Subspaces}
\citet{gur2018gradient} showed strong empirical evidence that the gradient dynamically converges to a very small subspace in various large-scale deep-learning scenarios. The subspace is spanned by a few top eigenvectors of the Hessian, and the dimension is equal to the number of data classes. This also indicates that the training trajectory of neural networks lies in a low-dimensional subspace, which is in line with the conclusion of \citet{li2022low}. Considering an illustrative example in Fig. \ref{fig:subspace}, the full parameter space contains three dimensions, but the training trajectory $\{\boldsymbol{\theta}_i^D\}_{i=0,..,t}$ only lies in a 2-dimensional subspace $\mathcal{S}$ spanned by $\boldsymbol{e}_1$ and $\boldsymbol{e}_2$. We call this subspace the intrinsic subspace because it has a minimal degree of freedom \citep{DBLP:conf/iclr/LiFLY18} for the objective function to reach the optimum. The aforementioned random subspace can be seen as a na\"ive estimation of $\mathcal{S}$.

We hypothesize that an intrinsic task-specific subspace exists for each downstream task when fine-tuning a PLM. Generally, it is intractable to search such an intrinsic task-specific subspace directly. However, if our hypothesis is true, the fine-tuning trajectory will lie in a low-dimensional subspace. Thus we can resort to the fine-tuning trajectory to obtain an approximation of the intrinsic task-specific subspace. Specifically, given a fine-tuning trajectory $\{\boldsymbol{\theta}_i^D\}_{i=0,..,t}$ of a PLM on a downstream task, we stack it into a matrix $\boldsymbol{W} \in \mathbb{R}^{t \times D}$, and apply Singular Value Decomposition (SVD) on it.
\begin{align}
  \boldsymbol{W} = \boldsymbol{U} \boldsymbol{\Sigma} \boldsymbol{V}^T,
  \label{eq:svd}
\end{align}
where $\boldsymbol{\Sigma} \in \mathbb{R}^{t \times t}$ is the singular value matrix, $\boldsymbol{U} \in \mathbb{R}^{t \times t}$ and $\boldsymbol{V} \in \mathbb{R}^{D \times t}$ are two real orthogonal matrices whose columns are left and right singular vectors, respectively\footnote{We assume $t \ll D$ and thus compact SVD is applied.}. It is worth noting that the columns of $\boldsymbol{V}$ are actually the principal directions of the given trajectory if zero empirical means of columns, and these directions constitute an orthonormal basis of the subspace in which the trajectory lies. Theoretically, a $(t-1)$-dimensional subspace needs only $t$ independent points to determine. We can regard this subspace as an approximation of the intrinsic task-specific subspace whose dimension is equal to the number of points in the trajectory. Thus, we can replace the random projection matrix $\boldsymbol{P}$ in Eq. (\ref{eq:subspace}) with $\boldsymbol{V}$ to re-parameterize the model.

\subsection{Fine-tuning in Intrinsic Task-specific Subspaces}
Given an approximated intrinsic task-specific subspace $\boldsymbol{V}$, we reformulate Eq. (\ref{eq:subspace}) by letting the model train in the subspace as follows.
\begin{align}
  \boldsymbol{\theta}^D = \boldsymbol{\theta}_0^D +  \boldsymbol{V} \boldsymbol{\theta}^t.
  \label{eq:subspace2}
\end{align}
In our early exploration, we can achieve good performance close to full fine-tuning by Eq. (\ref{eq:subspace2}). However, the performance is not stable, and sensitive to the initialization of $\boldsymbol{\theta}^t$. To solve this problem, we propose an ensemble-like method that combines multiple $\boldsymbol{\theta}^t$ of different initialization to reduce variance, which is as follows.
\begin{align}
  \boldsymbol{\theta}^D = \boldsymbol{\theta}_0^D +   \boldsymbol{V} \sum_{i=1}^h \frac{1}{h}  \boldsymbol{\theta}^{t(i)},
  \label{eq:subspace3}
\end{align}
where $h$ is the number of vectors to combine, and we set it as 16 in this paper. Note that although the ensemble increases the number of parameters to optimize, it does not change the instrinsic dimensionality of the subspace (i.e., the degree of freedom).

In the following experimental evaluation, we will investigate subspace fine-tuning in both transductive and inductive settings to verify our hypotheses. The former is to verify the existence of intrinsic task-specific subspaces when fine-tuning PLMs on the downstream tasks, and the effectiveness of our method to uncover the subspaces. The latter further examines how well the intrinsic task-specific subspaces can be transferred to other similar tasks.

\section{Experiment and Analysis}
\label{sec:experiment}
\subsection{Experimental Settings}

\begin{table*}[t!]
  \centering
  \begin{widetabular}{\textwidth}{lccccccccc} 
  \toprule
   & \textbf{CoLA} & \textbf{MRPC} & \textbf{SST-2} & \textbf{STS-B} & \textbf{QQP} & \textbf{MNLI} & \textbf{QNLI} & \textbf{RTE} & \textbf{Avg.} \\ \hline
  BERT-Full      & \underline{59.37} & \textbf{84.46} & \textbf{91.95} & \underline{89.08} & \textbf{91.07} & \textbf{83.39} & \textbf{90.77} & \underline{66.93} & \textbf{82.13} \\
  BERT-Freeze    & 27.52 & 69.66 & 88.81 & 78.35 & 84.48 & 71.55 & 81.61 & 56.46 & 69.81\\
  BERT-Random    & 37.89 & 70.78 & 89.47 & 81.41 & 85.86 & 72.91 & 83.38 & 58.63 & 72.54\\
  BERT-Intrinsic & \textbf{60.27} & \underline{84.31} & \underline{89.93} & \textbf{89.51} & \underline{89.73} & \underline{81.21} & \underline{87.73} & \textbf{67.00} & \underline{81.21}\\
  \hline

  RoBERTa-Full      & \underline{61.04} & \textbf{89.31} & \textbf{94.29} & \textbf{90.70} & \textbf{91.72} & \textbf{87.23} & \textbf{92.48} & \underline{76.68} & \textbf{85.43} \\
  RoBERTa-Freeze    & 0.00  & 68.38 & 85.32 & 15.69 & 82.81 & 71.16 & 79.11 & 53.86 & 57.04\\
  RoBERTa-Random    & 27.58 & 68.38 & 91.45 & 75.47 & 86.33 & 77.10 & 84.49 & 58.27 & 71.13\\
  RoBERTa-Intrinsic & \textbf{61.07} & \underline{87.21} & \underline{92.43} & \underline{89.43} & \underline{90.18} & \underline{85.53} & \underline{90.57} & \textbf{78.77} & \underline{84.40}\\
  \bottomrule
  \end{widetabular}
  \caption{Transductive intrinsic subspace fine-tuning on the GLUE benchmark. \textit{Full} denotes fine-tuning in the full-parameter space. \textit{Freeze} denotes fine-tuning with the encoder frozen. \textit{Random} denotes fine-tuning in a random subspace. \textit{Intrinsic} denotes fine-tuning in the intrinsic task-specific subspaces. The subspace dimension is set to 32 except MNLI is 64. The best results are marked in bold, and the second-best results are underlined.}

\label{table:transductive}
\end{table*}

\begin{figure*}[htpb]
  \begin{center}
    \includegraphics [width=\columnwidth]{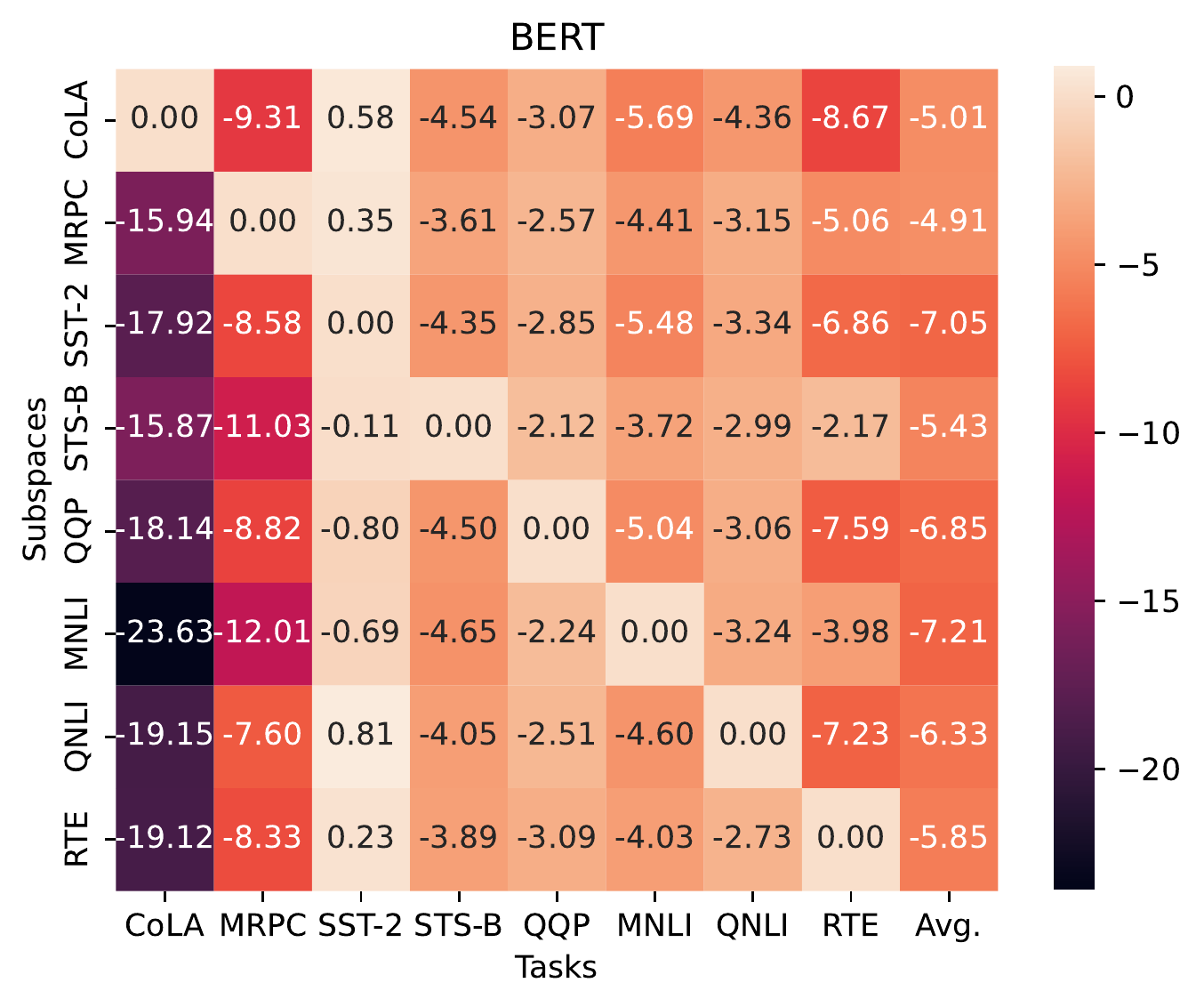} \hspace{0pt}
    \includegraphics [width=\columnwidth]{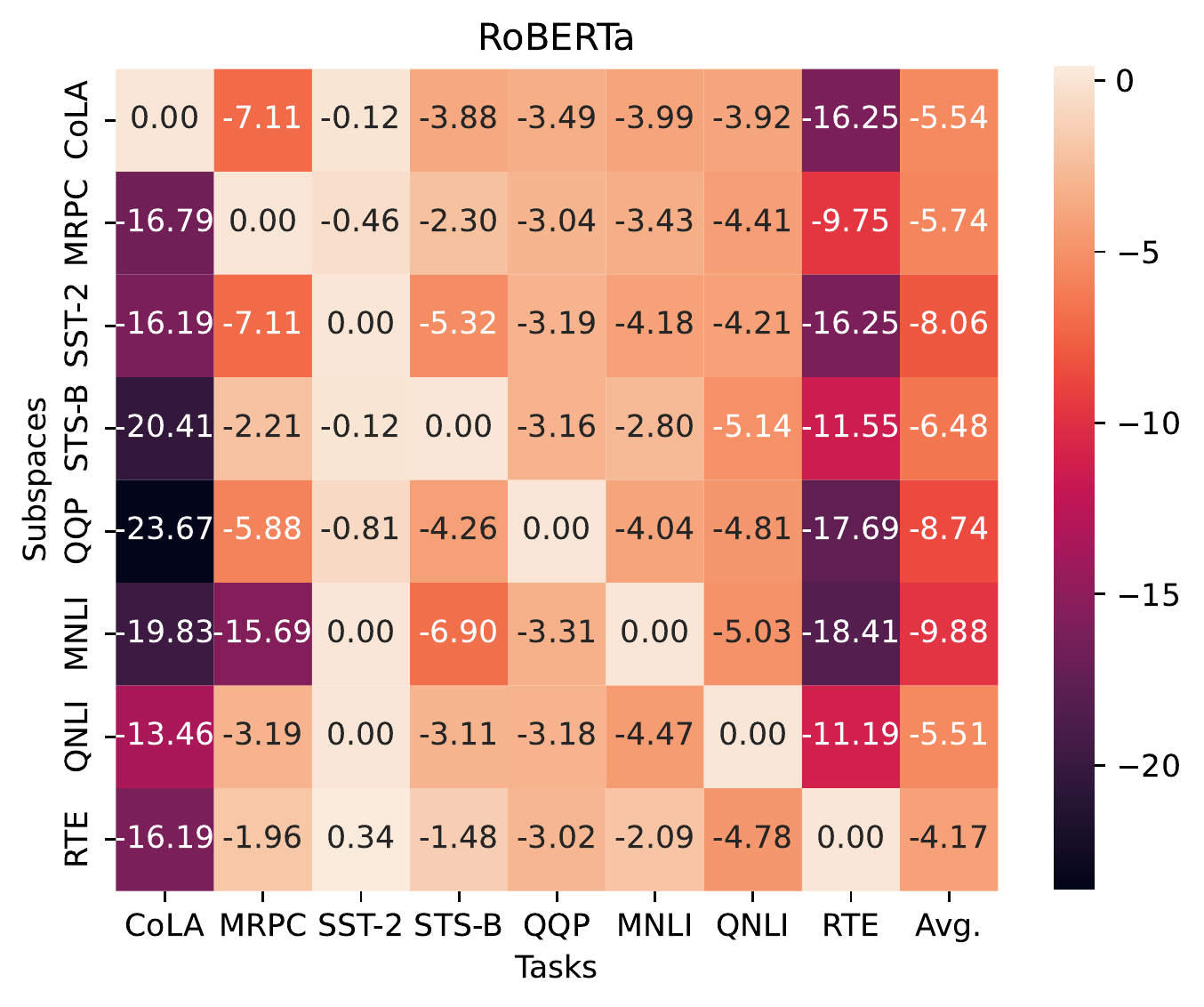}
  \end{center}
  \caption{Inductive intrinsic subspace fine-tuning on the GLUE benchmark. Columns are the target tasks to be fine-tuned, and rows are source tasks that provide the transferred subspaces. Numbers in cells are performance drop of fine-tuning target tasks with the subspaces provided by source tasks. The last column is the average of other columns. Note that the numbers cannot be compared across columns because they are in different metrics. }
  \label{fig:transfer}
\end{figure*}

\vspace{1mm}
\noindent \textbf{Datasets and models}. We evaluate the performance of the methods on the commonly used GLUE benchmark \citep{wang2018glue,DBLP:journals/tacl/WarstadtSB19,socher-etal-2013-recursive,dolan-brockett-2005-automatically,cer-etal-2017-semeval,williams-etal-2018-broad,rajpurkar-etal-2016-squad}. For evaluation metrics, we report the matched accuracy for MNLI, Matthew’s correlation for CoLA, Pearson correlation for STS-B, and accuracy for other tasks. We choose the publicly available pre-trained language models RoBERTa-base \citep{liu2020roberta} and BERT-base-cased \citep{devlin2019bert} for analysis. All experimental results are averaged over 5 runs of different seeds.

\vspace{1mm}
\noindent \textbf{Implementation details}. Our implementation is based on HuggingFace's Transformers toolkit \citep{wolf-etal-2020-transformers}. We first need to produce a set of fine-tuning trajectories of GLUE tasks for calculating projection matrices. We use the default script in the toolkit for fine-tuning, and save a checkpoint every epoch to obtain optimization trajectories. We set the trajectory length to 32 except for the MNLI dataset, which is set to 64 since it is the largest dataset and needs more parameters to fit. We flatten all parameters in an encoder layer into a wide vector, and then stack all vectors of different checkpoints into a matrix to perform SVD. We compute independent projection matrices for all layers, resulting in 12 projection matrices. For transductive subspace fine-tuning, the projection matrix is calculated from the same task, while for inductive subspace fine-tuning, it is calculated from other tasks. We only re-parameterize the encoder layers into the subspaces and leave the embedding layer and the last classification layer in their original parameter space. We freeze the initial model $\boldsymbol{\theta}_0^D$ and the projection matrix $\boldsymbol{V}$, and only tune the low-dimensional vector $\boldsymbol{\theta}^t$. We keep the learning rate of the embedding and classification layers unchanged and set the learning rate of $\boldsymbol{\theta}^t$ to 0.01.

\subsection{Transductive Intrinsic Subspace Fine-tuning}

Table \ref{table:transductive} summarizes the experimental results. We can see that freezing the encoder significantly degrades the model performance as it serves as a na\"ive baseline (Note that it implies fine-tuning in the null space, i.e., $\boldsymbol{V} \boldsymbol{\theta}^t=\boldsymbol{0}$, which brings no information to update the model). For intrinsic subspace fine-tuning, we can clearly see that it shows comparable performance to the full fine-tuning across all GLUE tasks and models. In contrast, random projection only yields a marginal improvement over the baseline, and significantly underperforms intrinsic subspace fine-tuning.

\begin{table*}[t!]
  \centering
  \begin{widetabular}{\textwidth}{lccccccccc} 
  \toprule
   & \textbf{CoLA} & \textbf{MRPC} & \textbf{SST-2} & \textbf{STS-B} & \textbf{QQP} & \textbf{MNLI} & \textbf{QNLI} & \textbf{RTE} & \textbf{Avg.} \\ \hline
   BERT-Full      & \underline{59.37} & \textbf{84.46} & \textbf{91.95} & \underline{89.08} & \underline{91.07} & \underline{83.39} & \underline{90.77} & {66.93} & \underline{82.13} \\
   BERT-Random   & 32.49 & 70.15 & 88.65 & 79.29 & 84.84 & 71.75 & 82.29 & 57.11 & 70.82  \\
   BERT-Zeroshot     & 35.35 & {78.09} & \underline{91.06} & {85.17} & {87.57} & {75.29} & {84.01} & \textbf{75.23} & 76.47  \\
   BERT-Unified      & \textbf{61.58}  & \underline{84.41}  & \underline{91.06}  & \textbf{89.71}  & \textbf{91.27}  & \textbf{83.85}  & \textbf{90.97}  & \underline{67.00}  & \textbf{82.48}\\
  \hline

  RoBERTa-Full      & \underline{61.04} & \textbf{89.31} & \textbf{94.29} & \underline{90.70} & \underline{91.72} & \textbf{87.23} & \textbf{92.48} & \underline{76.68} & \underline{85.43} \\
  RoBERTa-Random  & 0.00 & 68.38 & 89.47 & 27.60 & 84.51 & 73.16 & 82.10 & 54.44 & 59.96 \\
  RoBERTa-Zeroshot   & {32.93}  & {80.44}  & {90.60}  & {83.10}  & {87.12}  & {78.76}  & {84.46}  & {67.12}  & 75.57 \\
  RoBERTa-Unified    & \textbf{63.80} & \underline{89.12} & \underline{93.55} & \textbf{90.88} & \textbf{91.85} & \underline{87.20} & \underline{92.36} & \textbf{77.91} & \textbf{85.83} \\
  \bottomrule
  \end{widetabular}
  \caption{Intrinsic subspace fine-tuning in the unified task subspace. \textit{Random} denotes fine-tuning in a random subspace (dim=8). \textit{Zeroshot} denotes fine-tuning in the unified task subspace with itself removed (dim=7). \textit{Unified} denotes fine-tuning in the unified task subspace (dim=8).}
\label{table:unified}
\end{table*}

From these empirical results, we first conclude that PLMs can be re-parameterized and fine-tuned in some low-dimensional subspaces. Secondly, there exist some subspaces in which the PLMs can most effectively adapt to downstream tasks, and we can uncover these subspaces by finding the principal directions of fine-tuning trajectories in the full parameter space. This conclusion in turn suggests that fine-tuning of PLMs happens in tiny subspaces, which provides an explanation of the ease of adapting PLMs to downstream tasks.

\subsection{Inductive Intrinsic Subspace Fine-tuning}

Next, we conduct inductive intrinsic subspace fine-tuning to examine the transferability of the discovered subspaces. We generally follow the same training protocol as in the last section, except that we replace the projection matrices with the ones calculated from other tasks. 

We can observe the performance drop using transferred task subspaces in Fig. \ref{fig:transfer}. Generally, we can see that even though the models are fine-tuned in transferred subspaces, they still outperform the random subspace baseline, which suggests the transferability of intrinsic task-specific subspaces. 

The transferability of subspaces seems to correlate with the scale of the transferred task. For example, big datasets like SST-2, QQP, MNLI and QNLI underperform small datasets like CoLA, MRPC, STS-B, and RTE in providing subspaces. This is because the intrinsic task-specific subspaces of complex tasks have higher dimensions and need more parameters to estimate.

When comparing within one column, we can see significant difference between distinct subspaces used for fine-tuning one task. We assume similar tasks may have substantial subspace intersections and thus be easier to transfer. Still, this claim needs further analysis to confirm, we will leave it further study since transferability is not the main focus of this paper. In summary, we empirically show that the intrinsic task-specific subspace has a certain transferability.

\subsection{Unified Intrinsic Task Subspace}

\citet{qin2021exploring} showed that a unified low-dimensional intrinsic task subspace can be constructed by a multi-task prompt tuning method. In our case, we can also construct a unified subspace by stacking the fine-tuning trajectories of different tasks into a matrix, and applying SVD on it. Specifically, we sample one checkpoint for each task and gather them to calculate the unified subspace, which forms an 8-dimensional subspace. And we additionally calculate a zero-shot subspace of a task for comparison, which is calculated by excluding the checkpoint of this task. The results are given in Table \ref{table:unified}. We can see that the models can be effectively fine-tuned in the unified subspace. For the zero-shot setting, the model performance decreases significantly, but still outperforms the random baseline.

\begin{figure}[htpb]
  \begin{center}
    \includegraphics [width=\columnwidth]{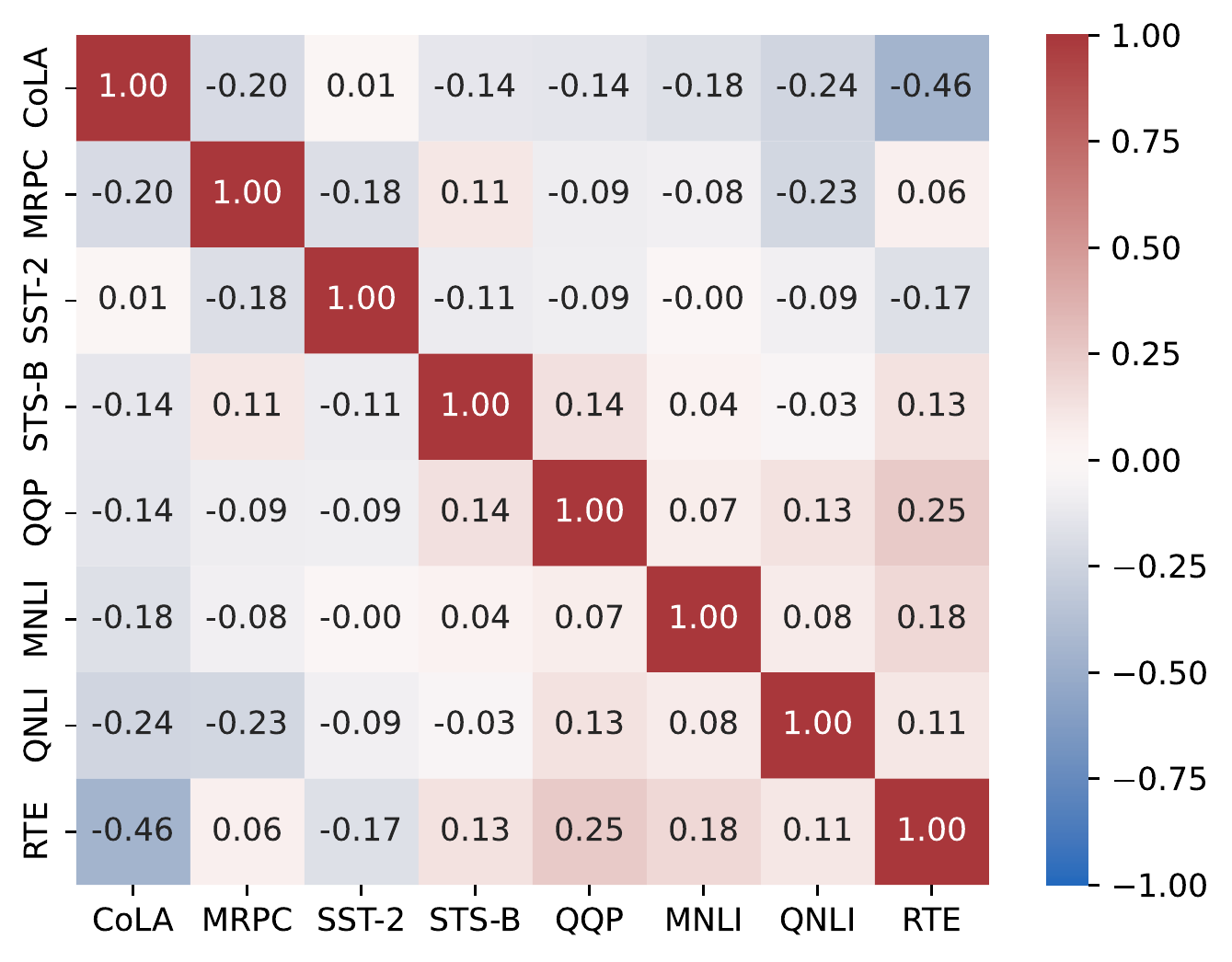}
  \end{center}
  \caption{The cosine similarities between the low-dimensional parameter vectors $\boldsymbol{\theta}^t$ of different tasks in the unified intrinsic task subspace. Similarities are averaged over layers and ensembles.}
  \label{fig:theta}
\end{figure}

\begin{figure*}[t!]
  \begin{center}
    \includegraphics [width=0.90\textwidth]{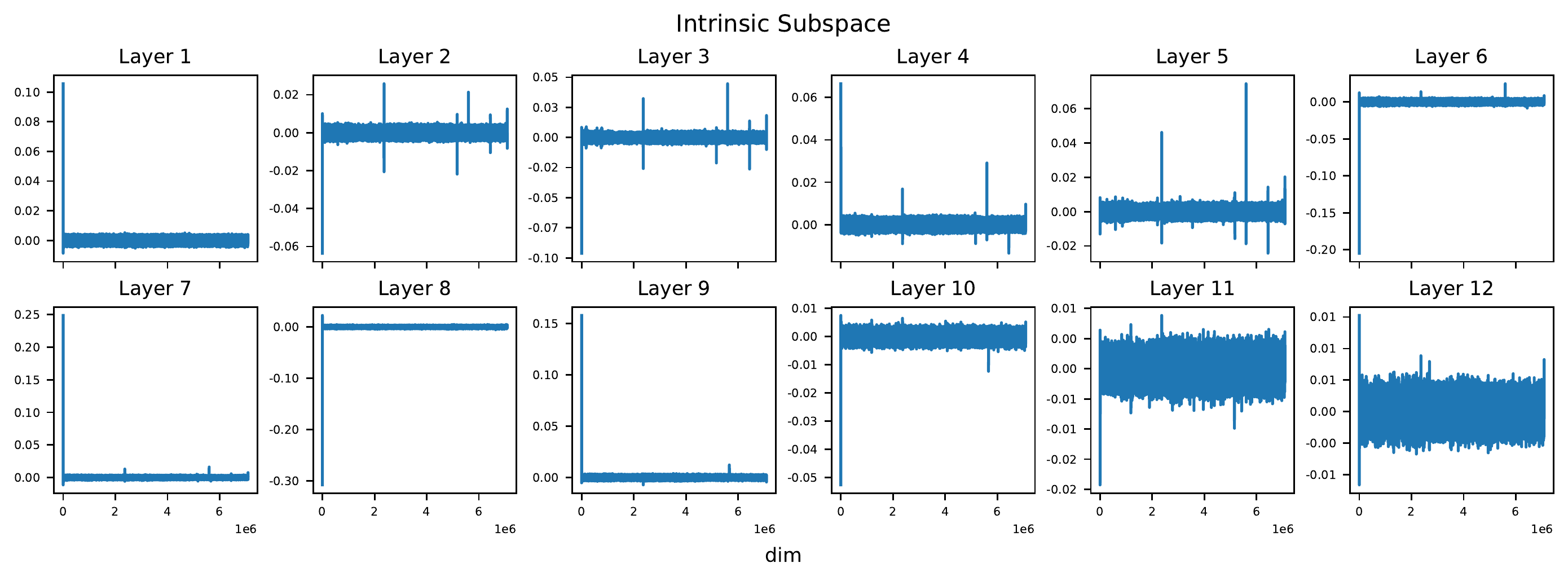} \hspace{0pt}
    \includegraphics [width=0.90\textwidth]{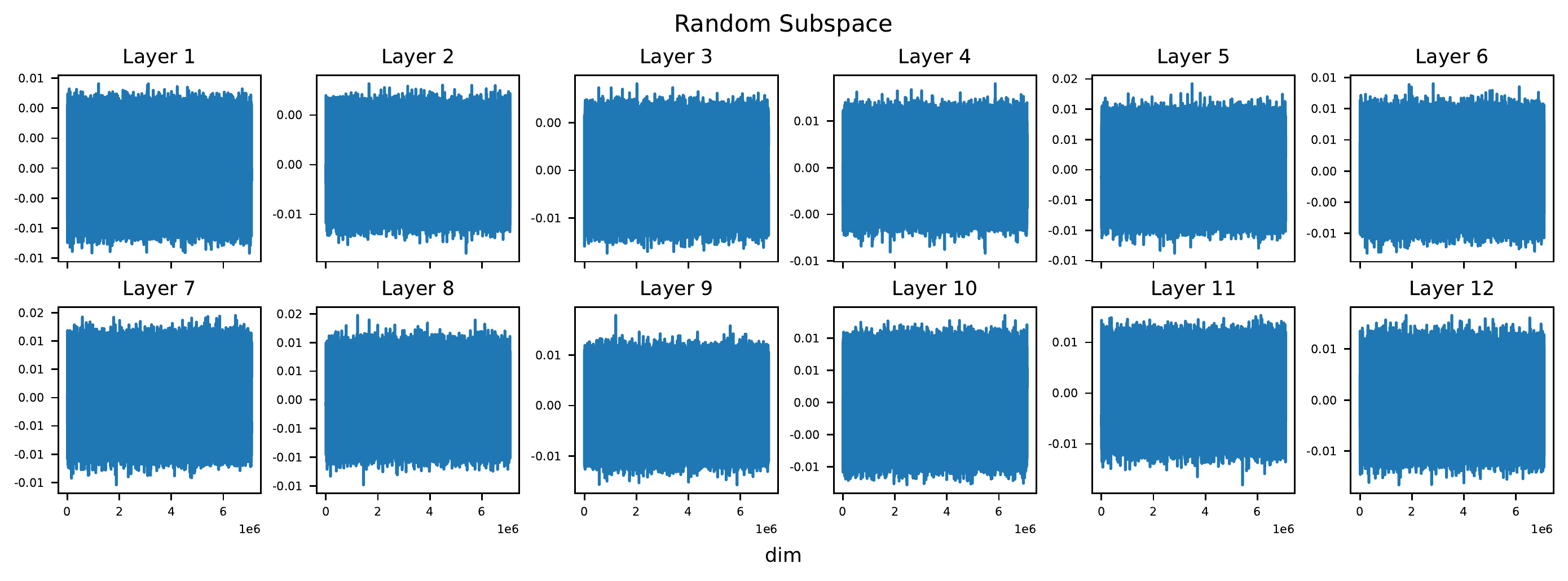}
  \end{center}
  \caption{Visualization of product of $\boldsymbol{V} \boldsymbol{\theta}^t$ using the BERT model to fine-tune in the intrinsic task-specific subspace (top) and a random subspace (bottom) on the MRPC dataset. Outlier dimensions appear in the intrinsic subspace but not in a random subspace.}
  \label{fig:outlier}
\end{figure*}

Next, we take the BERT model as an example and examine the low-dimensional parameter vector $\boldsymbol{\theta}^t$ learned within the unified intrinsic subspace. We calculate the cosine similarities between the $\boldsymbol{\theta}^t$ vectors corresponding to different tasks and present the results in Fig. \ref{fig:theta}. As shown in the figure, the cosine similarities between different tasks are significantly low, indicating that the unified intrinsic subspace contains disentangled knowledge distributed in different dimensions, and the low-dimensional parameter vector $\boldsymbol{\theta}^t$ serves as an (unnormalized) probability distribution to induce task-specific knowledge.

Based on these empirical findings, we conclude that a unified intrinsic task subspace is feasible and it contains disentangled knowledge. However, in-domain knowledge still plays a crucial role in forming the subspace as we can see that the zero-shot setting still has a large perform gap.

\subsection{Outlier Dimensions}

\begin{table*}[htpb!]
  \centering
  \begin{widetabular}{\textwidth}{lcccccccc} 
  \toprule
   & \textbf{CoLA} & \textbf{MRPC} & \textbf{SST-2} & \textbf{STS-B} & \textbf{QQP} & \textbf{MNLI} & \textbf{QNLI} & \textbf{RTE} \\ \hline
   BERT-Full      & 59.37 & 84.46 & 91.95 & 89.08 & 91.07 & 83.39 & 90.77 & 66.93 \\
   BERT-Random & 57.27 & 84.46 & 91.79 & 88.66 & 90.66 & 83.68 & 90.41 & 64.48 \\
   BERT-Outlier & \textbf{0.00} & \textbf{68.38} & \textbf{50.92} & \textbf{0.00} & \textbf{63.18} & \textbf{33.64} & \textbf{49.89} & \textbf{52.71} \\
  \hline
  RoBERTa-Full      & 61.04 & 89.31 & 94.29 & 90.70 & 91.72 & 87.23 & 92.48 & 76.68 \\
  RoBERTa-Random & 58.80 & 87.65 & 93.95 & 89.52 & 91.29 & 87.76 & 92.61 & 68.88  \\
  RoBERTa-Outlier & \textbf{0.00} & \textbf{70.49} & \textbf{50.92} & \textbf{28.05} & \textbf{63.67} & \textbf{36.15} & \textbf{49.89} & \textbf{52.71} \\
  \bottomrule
  \end{widetabular}
  \caption{Evaluation on the GLUE benchmark when the outlier dimensions are zeroed. The results with the most performance loss are marked in bold.}
\label{table:disable}
\end{table*}

\begin{table*}[t!]
  \centering
  \begin{widetabular}{\textwidth}{lll} 
  \toprule
  \textbf{Model component} & \textbf{Layer} & \textbf{\# of outliers each layer} \\ \hline
  attention.self.query.weight         & 1, 2, 3, 4, 6, 7, 8, 9, 10, 11, 12  & 3, 1, 1, 1, 4, 4, 8, 3, 3, 2, 4 \\
  attention.self.query.bias             & 1  & 1\\
  attention.self.key.bias               & 10, 11 & 2, 1\\
  attention.output.LayerNorm.weight    & 1, 2, 3, 4, 5, 6, 7, 9, 10, 11, 12 & 1, 2, 3, 5, 4, 1, 2, 4, 1, 3, 2 \\
  attention.output.LayerNorm.bias       & 1, 2, 3 & 1, 1, 1\\
  intermediate.dense.weight             & 1, 12 & 2, 1\\
  output.dense.weight       & 1, 2, 3, 4, 5, 6, 7, 8, 9, 10, 11 & 2, 6, 5, 4, 2, 4, 3, 2, 3, 4, 4
  \\
  output.LayerNorm.weight   & 5, 6, 7, 12 & 4, 1, 1, 3 \\
  \bottomrule
  \end{widetabular}
  \caption{Sampled outlier dimensions in the BERT model. The left column shows the model component containing outlier dimensions. The middle column shows the layer where the model component contains outlier dimensions. The right column shows the number of outlier dimensions in the corresponding layer.}
\label{table:outliers}
\end{table*}

We find that PLMs have a small number of outlier dimensions exhibiting abnormal spikes when fine-tuning in the intrinsic task-specific subspaces. We examine each dimension of the product of $\boldsymbol{V} \boldsymbol{\theta}^t$ and consider the dimension whose absolute value is greater than a threshold as outlier. Note that the product of $\boldsymbol{V} \boldsymbol{\theta}^t$ is the learned parameter update in the full parameter space and we re-parameterize the encoder of the PLM layer-wisely, thus it is a vector with the dimension equal to the number of all parameters of an encoder layer. 

It is important to note that the outlier dimension in our context is different from the previous studies \citep{DBLP:conf/acl/KovalevaKRR21,DBLP:conf/acl/LuoKM20,puccetti2022outliers}. Previous studies use the outlier dimension to refer to the output channel (768 dimensions for BERT-base). In our context, we flatten all parameters of a layer into a vector (7,087,872 dimensions for BERT-base). Then an outlier dimension refers to a specific parameter weight in the layer. We use the BERT model and MRPC dataset for illustration, and visualize the product of $\boldsymbol{V} \boldsymbol{\theta}^t$ in Fig. \ref{fig:outlier} to show the outlier patterns. As we can see from the figure, when fine-tuning in the intrinsic task-specific subspace, the outlier patterns exist in all layers. In contrast, these outlier patterns disappear when fine-tuning in a random subspace. This phenomenon is universal for different models and different datasets.

To investigate the effect of the outlier dimensions on the models, we disable them by setting them to zero and examine how this affects model performance. We first disable the top outlier dimension of each encoder layer and fine-tune the model in the full parameter space, which has almost no impact on model performance. This result is not surprising because disabling only one weight in a layer definitely has a negligible effect on the output than disabling an output channel as the previous studies do. We continue to disable more outlier dimensions, and these deviating at least $3\sigma$ from the mean are disabled. Approximately 0.3\% of encoder parameters are disabled. We also randomly sample and disable the same number of dimensions for comparison, and the results are shown in Table \ref{table:disable}. We can see that disabling outlier dimensions degrades the model performance significantly while disabling random dimensions does not.

Next, we qualitatively examine the positions in which the outlier dimensions emerge. We sample each layer's top 10 outlier dimensions and record their positions in Table \ref{table:outliers}. We can see that the outlier dimensions are ubiquitous in various model components. Then, we identify one outlier dimension $O1$ that consistently produces high-magnitude weights in almost all BERT layers. Furthermore, we find that there is a considerable overlap in the outlier dimensions of each layer, which suggests that these dimensions can propagate through layers.

Why do outlier dimensions emerge? Previous studies came up with several explanations like high-magnitude scaling factors \citep{DBLP:conf/acl/KovalevaKRR21}, LayerNorm and residual connection \cite{DBLP:conf/acl/LuoKM20}, and unbalanced token frequency \citep{puccetti2022outliers}. However, these explanations cannot apply to our case because the definitions of the outlier dimension are different. Recall that our approach to identifying outlier dimensions is actually examining re-parameterized parameter updates given the intrinsic task-specific subspace. The magnitude of the updates represents the importance of corresponding parameters with respect to solving the task. We have reason to believe that these dimensions play an important role in constituting the intrinsic subspace and are crucial to induce task-specific knowledge to adapt to downstream tasks.

\section{Conclusion}
In this paper, we claim that the fine-tuning of PLMs happens in tiny subspaces. To uncover such intrinsic task-specific subspaces, we exploit the fine-tuning trajectory to find its main direction. Our empirical experiments show that PLMs can effectively adapt to downstream tasks when re-parameterizing and training in the found subspaces, which well explains the ease of adapting PLMs to downstream tasks. Furthermore, we find outlier dimensions in PLMs during the subspace training. We consider that these dimensions are crucial to induce task-specific knowledge to downstream tasks. Still, we need further in-depth analysis to understand the reasons and impact of the emergence of outlier patterns.

\section*{Limitations}

Despite the insights obtained through our analysis, certain limitations persist, which we outline in this section.

With respect to the re-parameterization of parameters as presented in Eq. (3), we adopted the layer-wise setting as proposed by Aghajanyan et al. (2021) in order to alleviate memory and computational burdens. Nonetheless, such a setting restricts us to only identifying local subspaces, rather than discovering global subspaces within the entire parameter space of a pre-trained language model. The existence of a task-specific global subspace is yet to be ascertained. If such a subspace does exist, the correlation between this global subspace and the identified local subspaces needs to be explored in future research.

In terms of experimental settings, the evaluation tasks are limited to natural language understanding tasks, with a lack of natural language generation tasks. On model architecture, decoder-only (e.g., GPT) and encoder-decoder (e.g., T5) models are not included. On model scale, we use basic-size models rather than large ones due to limited computational resources. Consequently, the conclusions drawn in this study may not be applicable to the above situations.

The analysis presented in Section 4.5 demonstrates that pre-trained language models exhibit a small number of outlier dimensions when fine-tuning in the intrinsic task-specific subspaces. Although we have observed a significant decline in model performance when disabling these dimensions, the underlying mechanism responsible for the emergence of these outlier dimensions remains unclear.

\section*{Acknowlegments}
This work is supported by the Sichuan key research program (22ZDYF3388), Fundamental Research Funds for the Central Universities (ZYGX2019Z014),  National Natural Science Foundation of China (61976044, 52079026), Fok YingTong Education Foundation for Young Teachers in the Higher Education Institutions of China (161062), the Canada CIFAR AI Chair Program, and the Canada NSERC Discovery Grant (RGPIN-2021-03115).

\bibliography{mybib}

\appendix

\section{Appendix}

\subsection{Hyperparameters}

We first fine-tune the BERT and RoBERTa models for calculating projection matrices. We use the fine-tuning script in the Transformers toolkit\footnote{\url{https://github.com/huggingface/transformers/tree/main/examples/pytorch/text-classification}}. All hyperparameters remain default except for the number of epochs, which is set to 32 and 64 for the MNLI and all other tasks, respectively. For intrinsic subspace fine-tuning, the dimensionality of $\boldsymbol{\theta}^t$ is set to 32 and 64 for the MNLI and all other tasks, respectively. The learning rate of $\boldsymbol{\theta}^t$ is set to 0.01. The number of ensembles $h$ is set to 16. Other hyperparameter are the same as those in the script. All experimental results are averaged over 5 runs of different seeds. Each experiment is conducted on a single GeForce RTX 2080Ti GPU with environment of Pytorch 1.11.0 + CUDA 11.3.1.

\subsection{Ablation study}

We conduct an ablation experiment over the number of dimensions of the subspaces. The results are given in Table \ref{table:ablation1} and Table \ref{table:ablation2}. The performance increases as the number of dimensions increases.

\begin{table}[h!]
  \centering
  \begin{widetabular}{\columnwidth}{lccc} 
  \toprule
   Tasks & dim=8 & dim=16 & dim=32 \\ \hline
   \textbf{CoLA}  & 54.06 & 57.17 & \textbf{60.27} \\
  \textbf{MRPC}   & 75.05 & 77.94 & \textbf{84.31} \\
  \textbf{SST-2}  & 89.52 & \textbf{90.05} & 89.93 \\
  \textbf{STS-B}  & 87.95 & 89.02 & \textbf{89.51} \\
  \textbf{QQP}    & 87.61 & 89.12 & \textbf{89.73} \\
   \textbf{MNLI}  & 76.93 & 78.48 & \textbf{78.70} \\
    \textbf{QNLI} & 86.54 & 86.83 & \textbf{87.73} \\
     \textbf{RTE} & 65.41 & 66.07 & \textbf{67.00} \\
  \bottomrule
  \end{widetabular}
  \caption{Ablation study for the BERT model.}
\label{table:ablation1}
\end{table}

\begin{table}[h!]
  \centering
  \begin{widetabular}{\columnwidth}{lccc} 
  \toprule
   Tasks & dim=8 & dim=16 & dim=32 \\ \hline
   \textbf{CoLA}  & 58.04 & 60.27 & \textbf{61.07} \\
  \textbf{MRPC}   & 75.59 & 78.20 & \textbf{87.21} \\
  \textbf{SST-2}  & 91.93 & 92.34 & \textbf{92.43} \\
  \textbf{STS-B}  & 84.10 & 88.10 & \textbf{89.43} \\
  \textbf{QQP}    & 87.58 & 89.25 & \textbf{90.18} \\
   \textbf{MNLI}  & 79.96 & 81.77 & \textbf{82.32} \\
    \textbf{QNLI} & 89.35 & 89.14 & \textbf{90.57} \\
     \textbf{RTE} & 74.30 & 78.56 & \textbf{78.77} \\
  \bottomrule
  \end{widetabular}
  \caption{Ablation study for the RoBERTa model.}
\label{table:ablation2}
\end{table}

\end{document}